# *A deep learning perspective on Rubens' attribution*

Angie Afifi[1]*, Artur Kalimullin[2], Sergey Korchagin[3], Ilya Kudryashov[4]

**Abstract**

This study explores the use of deep learning for the authentication and attribution of paintings, focusing on the complex case of Peter Paul Rubens and his workshop. A convolutional neural network was trained on a curated dataset of verified and comparative artworks to identify micro-level stylistic features characteristic of the master's hand. The model achieved high classification accuracy and demonstrated the potential of computational analysis to complement traditional art historical expertise, offering new insights into authorship and workshop collaboration.

*Correspondence:
Angie Afifi
eafifiaf40@alumnes.ub.edu
[1]Faculty of Geography and History, Department of Art History, University of Barcelona, Spain
[2]Art Discovery, USA
[3]ORCID: 0000-0003-0678-8372
[4]ORCID: 0009-0009-1889-6232

## I. Introduction. The historical and cultural phenomenon of art forgery

The relevance of studying art forgery stems from their scale, their impact on the functioning of the art market and artistic institutions, as well as the fact that they affect fundamental categories in art history, such as authorship, style, provenance, and historical context. The history of art forgeries demonstrates how deeply the phenomenon of forgery is embedded in the institutional, market, and intellectual infrastructure of art. From antiquity to the present day, forgery has invariably accompanied the evolution of the concept of artistic value, taking various forms from devotional imitation to systematic commercial deception. In different historical contexts, artistic forgeries have fulfilled diverse functions, including sacred, rhetorical, and commercial. The names of artists whose work has suffered most from forgeries vividly illustrate the structural fractures of the artistic field and the vulnerability of expert mechanisms.

In the Middle Ages, under the dominance of ecclesiastical commissions, the primary targets of forgery were relics and icons. Already in the Renaissance era, with its cult of antiquity and the increasing importance of individual authorship, a new type of forgery emerged based on stylistic and ideological imitation, which was often perceived as an expression of reverence for the past and a demonstration of skill. Notably, at that time such forgery did not carry a negative connotation. A talented reproduction of antique models was seen as a sign of artistic merit rather than an attempt to deceive (Lenain, 2012; Scott, 2016). It was precisely through such a work, styled as an antique sculpture, *Sleeping Cupid*, that the young Michelangelo first attracted the attention of Lorenzo de' Medici. The work was subsequently artificially aged and sold as a genuine antique statue. Thus, it is during this period that the first documented cases of stylistic forgeries appear, receiving public approval and market resilience.

The establishment of the first auction houses in the 18th century, such as Sotheby's, Christie's, Bonhams, and Phillips, and the formation of public collections in the 19th century, were accompanied by a rapid growth in demand for Old Master paintings (Arnau, 1961). This demand, in turn, fostered the emergence of highly skilled forgers capable not only of imitating painterly techniques but also of manipulating provenance. Particularly vulnerable were artists whose studio organization involved serial production and the participation of assistants and apprentices. For example, the workshop of Lucas Cranach the Elder functioned almost like a manufacturing corporation. Dozens of versions of popular subjects were created there, often bearing the master's monogram even when the authorship belonged to apprentices (Heydenreich, 2007). This system continued to operate after the artist's death, under the direction of Cranach the Younger, significantly complicating the differentiation between originals, workshop productions, and later copies.

The work of Pieter Bruegel the Elder also gave rise to a prolonged tradition of imitative practices, continued by his sons and numerous followers (Ainsworth & Christiansen, 1998). Attribution challenges are further intensified in the case of Raphael, whose Roman workshop included a large number of assistants, among whom Giulio Romano executed major parts of monumental commissions (Ugail et al., 2023). The mass reproduction of etchings by Rembrandt and late Goya also created substantial difficulties in distinguishing original prints, later editions from original plates, and forgeries produced from falsified or third-party engravings (Spencer, 2004; Arnau, 1961).

All these examples demonstrate how the structural features of studio-based art production created preconditions for the emergence of forgeries and greatly complicated the attribution of works either to the master himself or to his workshop, followers, or forgers.

Another striking testimony to the difficulty of establishing authorship and the widespread presence of works with questionable attribution is the oeuvre of Rubens. This body of work is of particular interest in the context of the development of our algorithm. Possessing one of the largest and most organized workshops of the Baroque era, Rubens systematically engaged numerous students and assistants. It is known that compositions were often executed by them and only subsequently revised or approved by the artist himself. This significantly complicates the determination of the degree of the master's personal involvement in the creation of specific works, as well as the differentiation between autograph works, workshop productions, and later imitations. An additional complication is posed by the high market value of Rubens' works, which, combined with the organization of his workshop, provides especially fertile ground for the appearance of forgeries (Arnau, 1961). This, in turn, necessitates a comprehensive approach to authentication that includes the use of modern machine learning methods.

In light of these challenges, the 19th and 20th centuries saw the active development of new methods for examining authenticity, against the backdrop of growing interest in the issue. With advances in art history and conservation science, approaches to technical and stylistic analysis began to take shape, including radiography, infrared imaging, as well as improvements in methods of chemical analysis of pigments and binding media. These developments significantly increased the objectivity of authenticity assessments (Ragai, 2015). Nevertheless, legal and public scandals surrounding the exposure of well-known forgers such as Han van Meegeren, Wolfgang Beltracchi, Eric Hebborn, and Tom Keating demonstrate that even with the availability of new technical methods, experts often remained vulnerable due to an overreliance on traditional stylistic criteria, which frequently represent standardized and simplified images of a given era (Lenain, 2012).

In the 21st century, in the context of the digitization of art, the emergence of NFTs, and the rapid growth of the global art market, forgeries have taken on new forms and spread at unprecedented speed. These processes raise not only questions of legal and financial accountability but also fundamental issues of authenticity, authorship, and the status of the artwork in contemporary culture (Lindemann, 2006; Radermecker & Ginsburgh, 2023).

Meanwhile, forgeries in the field of modernist painting have reached considerable scale. This is due to the relative technological ease of execution, the wide range of available materials both historical and modern, similar in composition and visual characteristics, as well as insufficient documentation. These factors have complicated the application of scientific analysis and reduced its evidentiary value (Museum Ludwig, 2020). Works by artists such as Amedeo Modigliani are subject to mass forgery. Attempts to catalogue his legacy continue to be accompanied by institutional contradictions. Different catalogues raisonnés include or exclude the same works based on conflicting criteria (Jackson & Francis, 2025). A similar situation surrounds the work of Henri Rousseau, where no consensus exists regarding the authenticity of a number of pieces (Haskell, 2012).

The case of Giorgio de Chirico is particularly specific. He deliberately problematized the very notion of authenticity by reproducing his early compositions and often backdating them. This led to an institutional crisis of trust, to the extent that even genuine works by the artist are frequently met with skepticism by experts (Spencer, 2004).

The misattribution of the painting *Black Rectangle, Red Square* to Kazimir Malevich, its inclusion in the state collection of North Rhine-Westphalia, and subsequent exposure also highlighted systemic weaknesses in institutional expertise. The scale of the problem is further confirmed by the results of a scientific study conducted by the Ludwig Museum in 2020. Almost half of the examined objects, previously attributed to masters of Russian and Soviet modernism, turned out to be fakes. This movement remains one of the most forged in art history. Its vulnerability is due both to the high demand for works by its representatives and to its complex historical trajectory, marked by repression, archival loss, and a lack of reliable provenance (Akinsha, 2020).

Thus, issues of attribution and authentication remain pressing even for the largest art institutions. Nevertheless, the primary channel for the circulation of forgeries remains the art market, especially auction houses. Unlike museums, where examination can be carried out in a more controlled environment, auction structures operate under tight deadlines and legal obligations that require swift decisions while simultaneously guaranteeing authenticity. This often results in errors. Numerous cases involving the legacy of van Gogh are particularly telling, as his work continues to be the subject of sharp attributional disputes (Arnau, 1961). A number of works previously attributed to him, having passed through auctions, were later withdrawn from sale or excluded from the scholarly corpus of his oeuvre (De la Faille, 1980; Bailey, 2024).

Against this background, an institutional crisis of trust is becoming increasingly apparent. Given the growing legal responsibility and the ambiguity of the consequences of expert opinions, a decline in activity is observed among specialists who previously played a key role in the process of establishing authorship. Refraining from public judgment has become a common practice (Spencer, 2004). This has a negative impact on the stability of the art market, undermines the authority of museums and academic institutions, and ultimately contributes to the further proliferation of forgeries.

In these circumstances, the need for new analytical tools becomes evident. Such tools must be capable of complementing both traditional connoisseurship and physico-chemical methods of investigating artworks. One such tool is machine learning, which shows great potential in attribution tasks. It is particularly promising in cases involving recurring stylistic motifs and features of painterly technique on a micro-level. These features are imperceptible to the human eye but detectable through computational methods.

This article is dedicated to the testing and art-historical interpretation of an innovative tool aimed at verifying the authenticity of paintings using machine learning algorithms. We propose to consider it not as a replacement for existing approaches but as an additional research method capable of identifying meaningful visual patterns and strengthening the argumentation in expert attribution.

The main contributions of this article are as follows:

- A literature review on the history of art forgeries, with a particular focus on Rubens
- The creation of a high-confidence dataset of artworks with clearly verified authenticity
- The development of a neural network-based method for forgery detection
- The creation of a new tool for art historians to support analytical work
- Preliminary conclusions regarding the attribution of disputed paintings: *Samson and Delilah* (ca. 1609–1611) and *Head of a Young Man* (1601–1602)

Thus, we aim to demonstrate how art history and new technologies can interact to form a more accurate and multidimensional approach to the issue of artwork authentication.

**II. The place of art historical expertise in contemporary forgery analysis**

Despite the fact that art historical expertise remains one of the fundamental elements in the history of art, central to the understanding of form, meaning, and value of a work of art, claims to its objectivity are increasingly called into question. For several decades, the traditional expert approach has been criticized both epistemologically and practically (Ebitz, 1988). However, even in the absence of "scientific" objectivity in the strict sense, it would be incorrect to speak of complete subjectivity and randomness in expert practice. There exists a range of patterns, recurring methodological strategies, and principles upon which this process is often based.

In a broad sense, art historical expertise represents the ability to "recognize" and "evaluate" a work of art based on visual and sensory perception. This process involves not only the identification of external characteristics but also a deep immersion into the particularities of artistic language, context, and technique. Visual analysis, as the foundation of expert evaluation, relies on developed visual perception, experience, and constant cross-checking of one's own observations (Gombrich, 1960; Ebitz, 1988; Cole, 2016). Repeated return to the work, comparison with other samples, and adjustment of conclusions are integral to expert practice.

Moreover, a central role is played by comparison of the analyzed work with securely attributed pieces by the artist, their workshop, or school, as well as iconographic and iconological analysis (Gombrich, 1960) and the ability to assess the quality of execution with

precision – a skill that is both a tool and a goal of expertise (Ebitz, 1988). However, visual analysis cannot be limited to superficial stylistic or iconographic description. It must include an understanding of optical, perceptual, and technical patterns underlying the artistic production of a particular period. The ability to recognize how a work operates with light, constructs perspective, or uses color becomes a key tool in detecting stylistic or technological inconsistencies that may signal forgery (Kemp, 1990).

The expert also considers issues of provenance, the origin of materials, studio organization, the degree of assistants' involvement, and other "invisible" parameters shaping the work in its historical context. Identifying material, technical, and documentary features that diverge from the author's body of work requires not only visual skills but also interdisciplinary training and analytical thinking (Ebitz, 1988; Cole, 2016).

Furthermore, a significant role is played by the systematic analysis of numerous visual markers, the so-called pictology, which involves decomposing the work into hundreds of characteristics such as line, form, rhythm, texture, brushstroke technique, color palette, treatment of light and shadow, choice of canvas, and compositional principles (Van Dantzig, 1973; Spencer, 2004). These markers may be conditionally divided into constructive, governed by laws of perception and composition, and expressive, reflecting the inner world of the artist.

Nevertheless, a legitimate question arises: to what extent do these methods bring us closer to objectivity? And if expert practice is indeed built upon visual sensitivity, memory, and accumulated experience, can it be regarded as a scientific procedure in the strict sense?

Since visual analysis largely relies on subjective perception, it is susceptible to the influence of numerous factors, from intuition and biased expectations to dominant cultural and institutional narratives.

Intuition, often praised as an inherent quality of an experienced expert, can simultaneously be a source of accurate recognitions and a cause of errors. We too often rely on stylistic or conceptual generalizations, and interpretation of visual markers can vary depending on context. Moreover, perception itself is not stable: it is subject to change under the influence of prior knowledge, professional background, and institutional pressure (Ebitz, 1988; Jones, 1990).

Art historical expertise is always conducted within a specific context, shaped by a body of previous and current attributions accepted within academic and museum communities, the art market, and the collector environment. As Gombrich (1960) and Cole (2016) note, truths in art history are rarely achieved through a rigorously calibrated procedure. On the contrary, they are formed through the complex interplay of subjectivity, authority, observation, logical reasoning, intuition, and trust within the professional community. This casts doubt on the illusion of full objectivity, especially in such informalized practices as authenticity verification or authorship determination.

Thus, even the most experienced experts are not immune to the influence of the human factor: their perception is structured by prior hypotheses, theoretical frameworks, and expectations. At the same time, the human factor can play both a positive role, facilitating the recognition of complex patterns through professional intuition, and a negative one, leading to bias and distorted interpretation.

Moreover, each new judgment is born within a particular cultural paradigm, where new attributions are considered convincing only when they logically fit into existing conceptions of the artist's creative development, environment, and the artistic period as a whole. Otherwise, such attribution is typically rejected (Ebitz, 1988).

Consequently, despite its significance, traditional art historical expertise is not without internal limitations. Yet even a comprehensive approach to expertise, including physico-chemical methods of analysis, is often insufficient. Increasingly sophisticated forgeries created using the same materials and tools as authentic works often become indistinguishable even by laboratory methods. In addition, physico-chemical research, despite its high informational value, can be costly and, in some cases, invasive, which limits its application (Ragai, 2015). This raises the issue of the need for innovative methods of visual recognition. Machine learning technologies are of particular value here, as they allow the detection of patterns and deviations in works of art at a level inaccessible to human perception (Ebitz, 1988; Jones, 1990; Spencer, 2004).

Algorithms, largely free from emotional and cognitive bias, are trained on extensive databases of high-quality images and metadata, enabling them to conduct not only quantitative but also visual analysis with a high degree of accuracy (Scott, 2016). The machine is capable of capturing micro-details, such as nuances of brushstroke, surface texture, micro-deviations in form and rhythm, which are difficult or impossible to perceive with the naked eye. This increases the likelihood of identifying anomalies and recurring patterns characteristic of either forgeries or authentic works.

These systems do not function in isolation. Their operation is most meaningful in conjunction with art historical expertise and other methods of analysis. Artificial intelligence acts as a supporting tool, processing often the same visual material as the expert, but relying on strictly analytical, quantitatively expressed, and systematically identified patterns. Thus, machine analysis does not replace visual judgment but enhances it, offering new forms of verification and improved reliability of attribution (Opperman, 1990, Charney, 2015).

### III. Problems of attribution in the context of Rubens' workshop

The issue of attribution in the case of Rubens occupies a central position in Rubens scholarship and remains one of the most debated problems in early modern art history (Van Hout, 2020). Already in the nineteenth century, we find the first significant attempts to systematize Rubens as a historical figure and to reflect on his workshop as a distinctive phenomenon within Baroque culture (Burckhardt, 1898). These early efforts, however, were often descriptive and rooted in broader cultural history rather than in technical or documentary analysis. It was only in the twentieth century that the study of Rubens' workshop developed into a fully professional field of research, thanks above all to the pioneering work of Ludwig Burchard and the subsequent publication of the Corpus Rubenianum Ludwig Burchard. From that moment, the question of distinguishing between the master's hand and that of his assistants became one of the central tasks of Rubens studies, shaping generations of art historians and continuing to structure debates up to the present (Balis, 1986; Vlieghe, 1972-73, 1987; Judson, 2000, Van Hout, 2020).

From the mid-twentieth century onwards, research began to stress the continuum of authorship in Rubens' studio, suggesting that modern categories of originality do not correspond neatly to seventeenth-century practices (Vlieghe, 1972-73). More recent studies have combined stylistic and iconographic analysis with technical and physico-chemical examinations to assess degrees of authorship and intervention. Even so, experts repeatedly note that the structural ambiguity of Rubens' oeuvre makes definitive answers rare, and that attribution must remain a field of ongoing interpretation (Van Hout, 2020).

The fundamental challenge lies in the collective character of Rubens' artistic production, carried out within a complex and prolific workshop in which assistants frequently executed substantial portions of paintings based on his designs, while he reserved for himself the invention, the most critical passages, or the final corrections. Scholars have long debated how such works should be classified: should they be considered autograph if the invention originates with Rubens, even when much of the execution belongs to another hand? Or should the category of workshop production be applied strictly to those cases where assistants dominated the surface of the canvas? The problem becomes even more intricate in cases of intermediate authorship, where Rubens collaborated closely with gifted pupils such as van Dyck or Jordaens. These collaborations produced hybrid works in which multiple hands are interwoven, but it remains difficult to identify with certainty which particular assistant contributed (Balis, 1986; Van Hout, 2020).

At the same time, workshop practice generated fertile ground for the proliferation of replicas, copies, and later imitations. Some works are faithful studio repetitions of lost prototypes, authorized and supervised by Rubens himself, while others are looser copies by students or later imitators. The result is a dense and overlapping corpus of images, where even experienced connoisseurs often disagree on classification. In fact, the very abundance of workshop material has ensured that attribution controversies are not marginal but form the core of Rubens studies.

In recent decades, the question has also been re-contextualized by technological and theoretical shifts. Technical art history has revealed how much can be learned from underdrawings, paint layers, and workshop processes invisible to the naked eye, while the digital age has opened new possibilities for computational approaches, including the application of machine learning to stylistic micro-markers. Yet even these innovations, while promising, underscore rather than resolve the central paradox: Rubens' workshop was deliberately designed to blur the boundaries of authorship, creating a collaborative model of production that resists the clear-cut categories demanded by modern connoisseurship and the art market alike. For this reason, the study of Rubens remains a paradigmatic case in the wider debate on authorship and authenticity in early modern art, a problem not only of attribution but also of cultural history.

**IV. Dataset formation**

The main goal of our research was to create and train a machine learning model to support the authentication of artworks by specific artists. For this purpose, a comprehensive dataset was formed containing both authenticated works of the selected artist and stylistically similar, yet clearly unauthentic pieces. The dataset consisted of high-quality images of the paintings, each standardized to a physical resolution of approximately 5 pixels per millimeter.

The dataset formation process involved two stages. First, a subset of artworks with undisputed authorship was selected and labeled as Certainty 1. Next, an art historian curated so-called negative examples - works by other artists, mostly contemporaries, whose style, use of color composition and level of expression resembled that of the main artist. In the case of Rubens, for example, this set included works by van Dyck and Jordaens, as well as a few paintings attributed to one of them or to Rubens, but with uncertain authorship. To maintain balance, we aimed to keep the number of artworks in each category roughly equal. Each image also passed through an automatic quality control step, which included resolution filtrating (e.g., pixels per millimeter) and detection of digitalization artifacts like noise, glare, reflections, and perspective distortion. This helped ensure visual consistency and minimized the influence of technical noise on the model's output.

The collected data were divided into three independent datasets: training, validation and test. This separation is essential for proper training and evaluation. The training set is used to teach the model to recognize the stylistic patterns of the target artist. The validation set allows for intermediate evaluation during training and helps in tuning model parameters. Finally, the test set is used for a final assessment of the model's ability to correctly classify artworks it has never seen before.

The artworks were split in an 80-10-10 ratio: the majority were allocated to the training dataset, with the remaining items evenly distributed between the validation and test sets. Class balance was maintained across all three subsets, ensuring that both positive and negative examples were present in each. To achieve comparable data volumes between classes across the datasets, we accounted not only for the number of whole artworks, but also for the total number of image fragments (tiles) into which they were later divided. This approach helped to ensure an approximately equal number of tiles per class and per dataset.

After the formation of the three datasets, all images were divided into small, fixed-size tiles of 512 by 512 pixels. This step is necessary because machine learning algorithms require a standardized input format. Since paintings typically vary in size and aspect ratio, tiling them into uniform fragments enables full spatial coverage of the image surface (although not of the entire composition, which is relevant for authenticity assessment and will be discussed below).

A detailed breakdown of artworks across the dataset is presented in Table 1:

| Dataset | No. of non-Rubens Works | No. of Rubens Works | No. of all Works | No. of non-Rubens tiles | No. of Rubens tiles | No. of all tiles |
|---|---|---|---|---|---|---|
| Train | 45 | 68 | 113 | 4783 | 5125 | 9908 |
| Val | 6 | 8 | 14 | 616 | 625 | 1241 |
| Test | 5 | 9 | 14 | 616 | 620 | 1236 |
| **Total** | 56 | 85 | 141 | 6015 | 6370 | 12385 |

Table 1. Distribution of artworks and cut-off tiles across datasets.

While art historians typically judge authenticity based on both the composition as a whole and its stylistic unity as well as fine details, in our study, we deliberately focused on the analysis of small-scale elements such as brushstroke characteristics, surface texture, and subtle stylistic features. Our underlying assumption is that such micro-level details, often too imperceptible to the human eye, can serve as objective indicators of authorship and significantly contribute to the precision of machine-based authentication.

Thus, the effectiveness of our system depended not only on the underlying algorithm, but also on the meticulous preparation of the dataset. A key factor was the balanced distribution of both authentic artworks and works by other artists that were stylistically and chronologically comparable, yet unrelated to the target authorship. This approach enabled the model to learn how to distinguish subtle, hard-to-detect elements of an artist's visual style from stylistic features that appear similar but originate from different sources.

**V. Tile analysis algorithm.**

To solve this task, we employed a machine learning algorithm approach based on a convolutional neural network (CNN) - a class of algorithms particularly effective for processing visual data. Specifically, we used the **MobileNetV4HybridLarge** architecture, a modern and efficient model widely applied in visual classification tasks. Its ability to balance analysis quality with computational efficiency makes it especially appropriate for use cases involving large-scale data under limited resource constraints (Qin et al., 2024).

To enhance both reliability and accuracy, we adopted an **ensemble of five independently trained neural networks** rather than relying on a single model. Each network made its own prediction for every tile, and their outputs were aggregated through an accumulation process that combines predictions from multiple models to strengthen the overall decision confidence. This ensemble strategy reduces the risk of individual model errors and yields more stable results, similar to how multiple art historian experts, independently analyzing a painting, may arrive at a more accurate consensus than a single expert.

Each model was trained for 200 epochs on the training dataset using the **Adam optimizer**, which automatically adapts the learning rate to improve the stability and speed of the optimization process. **Binary cross-enthropy (BCELoss)** was employed as the loss function, allowing the model to assess how confidently it can distinguish authentic artworks from non-authentic ones.

Special emphasis was placed on **data augmentation**, techniques that artificially increase image diversity and improve the model's robustness to stylistic and technical variations. These included random cropping, rotations, horizontal flips, controlled noise and distortion, adjustments in contrast and color balance, as well as more advanced transformations such as perspective shifts and elastic deformations. Such augmentations simulate the wide range of visual conditions under which artworks may be created or digitized, helping the model focus on meaningful stylistic features rather than superficial visual differences.

To prevent overfitting, when the model memorizes specific examples rather than learning generalizable patterns, a separate validation set was used during the training. This dataset was excluded from the direct learning and served as an objective measure of the model's ability to recognize authorship-related patterns. In addition to monitoring generalization performance, the validation set was also used for model tuning, such as determining the **optimal classification threshold**. The threshold defines the minimum predicted probability above which an example is classified as belonging to a positive class. This helps to convert the model's continuous output (e.g. 0.82) into a binary decision, "yes" or "no". Proper threshold selection allows balancing sensitivity and specificity of the system.

Finally, after training and validation-based tuning, the model was evaluated on the test set, which contained previously unseen data. This final evaluation provides an objective measure of the model's ability to identify authorship patterns on new, unfamiliar inputs.

## VI. Performance metrics

To evaluate how well the model performs its task, it is essential to use quantitative evaluation methods known as **performance metrics**. These metrics allow for an objective comparison between models, facilitate tracking of progress during training, and support informed decisions regarding parameter tuning and model improvements.

In our study, we employed several such metrics to assess the performance of the trained algorithm. One of the key indicators was **accuracy** - the proportion of instances for which the model correctly identified whether a given example belonged to the studied artist.

It is important to distinguish between two levels of accuracy evaluation:

- **Tile-level accuracy** reflects the model's ability to correctly classify individual fragments (tiles) of a painting. This metric captures the model's local sensitivity - its capacity to detect authorship-specific features within small regions of the artwork.
- **Image-level accuracy**, on the other hand, indicates how accurately the model classifies an entire painting, taking into account the predictions for all its constituent tiles. This is typically calculated by averaging the outputs across all tiles within an image.

It is important to note that the model is trained *exclusively* in tiles and never sees the full painting in its entirety. Therefore, high image-level accuracy does not imply that the model has learned to interpret global composition. Instead, it aggregates local signals derived from individual regions of the painting and makes a final decision based on their collective output.

In addition, we evaluated the ensemble's confidence in its predictions. Each neural network in the ensemble independently predicts the class of a given tile, and their outputs are compared. High agreement among the models indicates high ensemble confidence. Conversely, significant disagreement - reflected by high variance among outputs - can highlight tiles that require more cautious interpretation or further examination.

To complement these quantitative metrics, we also introduce a set of visualization tools designed to reveal how the ensemble arrives at its decisions. While not metrics in the strict numerical sense, these visualizations serve as qualitative evaluators of model behavior. They allow us to inspect where the ensemble expresses confidence, where its predictions diverge, and local tile-level evidence accumulates into an image-level judgement. In the next section, we integrate these visual tools into the presentation of our results, demonstrating how uncertainty and confidence maps help interpret model performance beyond raw values.

**VII. Interpretation of model outputs through an art historical lens**

The best results were achieved on the test dataset, with an image-level classification accuracy of approximately 85%, indicating that the ensemble was indeed able to reliably distinguish between stylistically similar works by different artists.

However, numerical accuracy alone does not reveal why the model succeeds or struggles on individual paintings. To interpret its decisions more transparently, we use complementary visualization techniques that highlight different aspects of the ensemble's behavior. The visualizations shown in the following figures illustrate the painting with two complementary overlays:

- The first visualization presents the **uncertainty map**. Each tile corresponds to a small fragment of the painting and its color intensity represents the degree of disagreement among the models within the ensemble. The more saturated the red color, the higher the disagreement. These strongly red-colored areas indicate regions where the ensemble had the most difficulty reaching a consensus. Pale or nearly transparent regions with low variance reflect high agreement, meaning all models predicted the same class.

- The second visualization displays the **confidence map**. This representation also depicts the model's perception of each tile but with a focus on its final confidence in classification. Highly saturated green regions correspond to high-confidence predictions above the chosen threshold, while highly saturated red regions correspond to high-confidence predictions below it. The more intense the color - whether green or red - the more certain the ensemble is about its decision. Tiles with faint or semi-transparent regions correspond to uncertain zones, where the confidence values are close to the threshold. In these regions, the ensemble has difficulty committing to either class, indicating that individual models are conflicting. Such low-confidence areas often (though not always) overlap with high-variance red zones in the uncertainty map, highlighting zones where disagreement among the networks makes classification more challenging.

To demonstrate the principles of the model's operation, we selected episodes analyzing both authentic works by Rubens and pieces stylistically reminiscent of his oeuvre but created by other artists. These examples clearly illustrate how the algorithm interprets the stylistic features it has internalized during training and detects deviations from them. This contrast helps to reveal both the potential and the limitations of the system in authenticity assessment.

*Aurora abducting Cephalus* **(ca. 1636-7)**

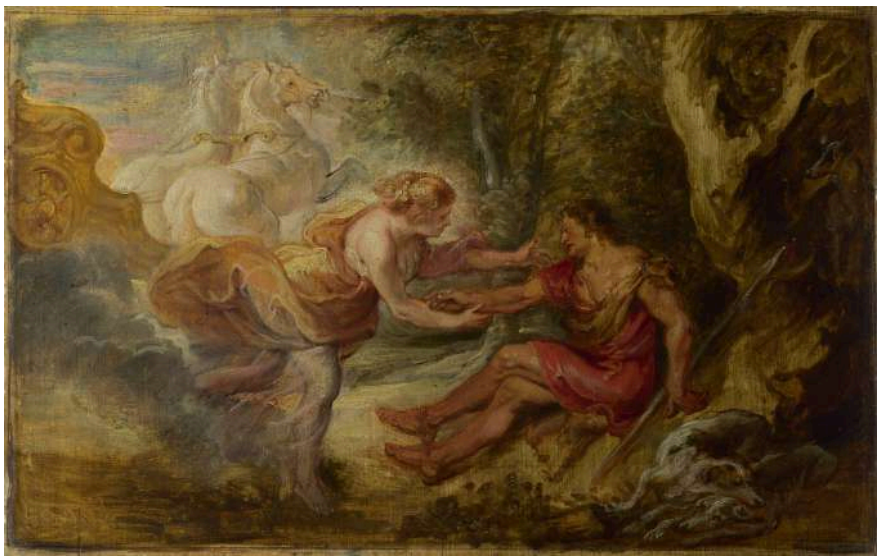

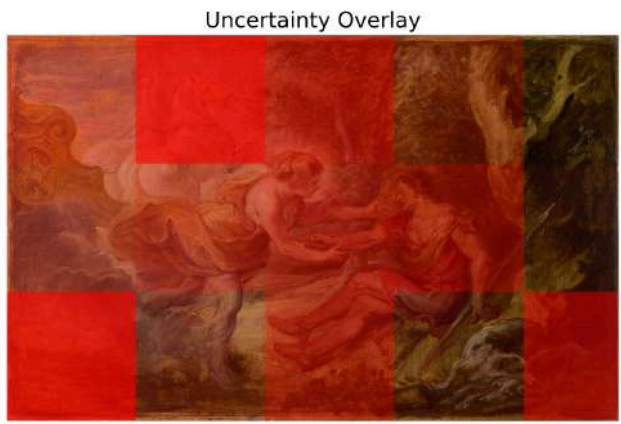
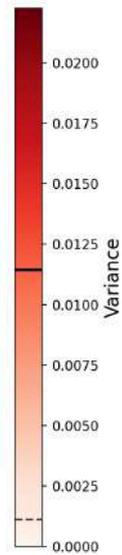

This work was recognized by the model as authentic Rubens, although the model's confidence level is relatively low: the average probability for the image is 0.6719, just above the classification threshold of 0.6080, and 13 out of 20 fragments were classified as consistent with Rubens' style.

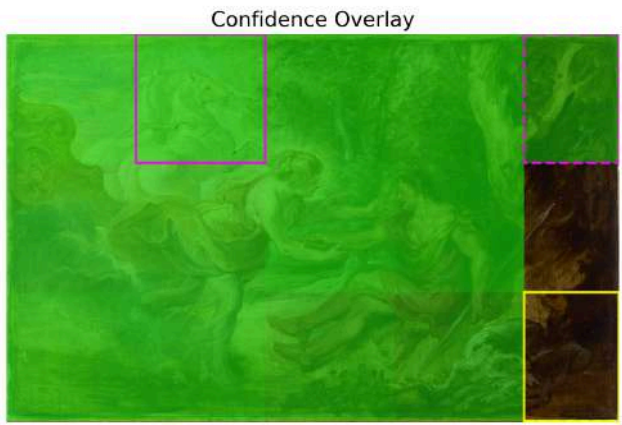
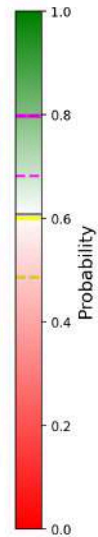

This indicates that certain areas of the canvas raise doubts for the algorithm, which is clearly visible on the Uncertainty map, where highly saturated regions correspond to strong disagreement within the ensemble. On the Confidence map, the fragments in the lower right part of the painting appear only faintly red and are nearly transparent. It reflects confidence values lying very close to the decision threshold. In such cases, the ensemble does not commit strongly to either class, suggesting that these tiles contain too little information, most likely due to cropping or the presence of visually ambiguous edge regions.

A similar effect can be seen in the upper right corner, where a green tile is marked as the one with the lowest ensemble probability, again indicating that insufficient visual context prevents the model from forming a confident judgement. These tiles are not diagnostically meaningful. Instead, they highlight the limitations of the tile-based classification in peripheral, incomplete, or low-detail areas.

From an art historical perspective, such uncertainty can be explained by stylistic deviations typical of this work, as we are not dealing with a finished composition but rather with an oil sketch, a preparatory reference piece. Its exploratory, experimental nature is evident in the less refined details and generalized forms. The model, trained primarily on mature, completed works by Rubens, encounters difficulties when analyzing such preparatory studies.

Particularly noteworthy is the area enclosed by a solid purple outline, which depicts two merging horse figures. This fragment was initially marked by the model as questionable and highlighted in red on the Uncertainty map, but during ensemble aggregation, it was ultimately classified as confidently authentic, as seen on the confidence panel. This shift in classification is especially telling in the context of Rubens' work, where the horse motif varies widely — from battle and mythological scenes to pastoral compositions — resulting in significant stylistic variability that can challenge automated classification. Nevertheless, despite deviating from standard templates, the model's final confidence in both this fragment and the overall composition indicates its ability to recognize the artist's manner even in visually ambiguous or stylistically borderline cases.

*Samson and Delilah* **(ca. 1609-11)**

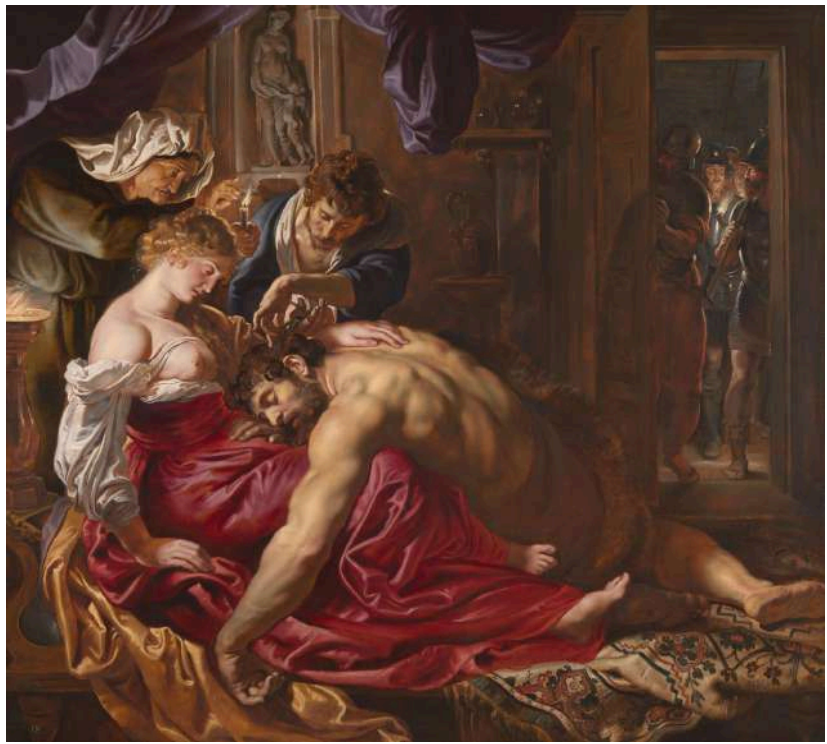

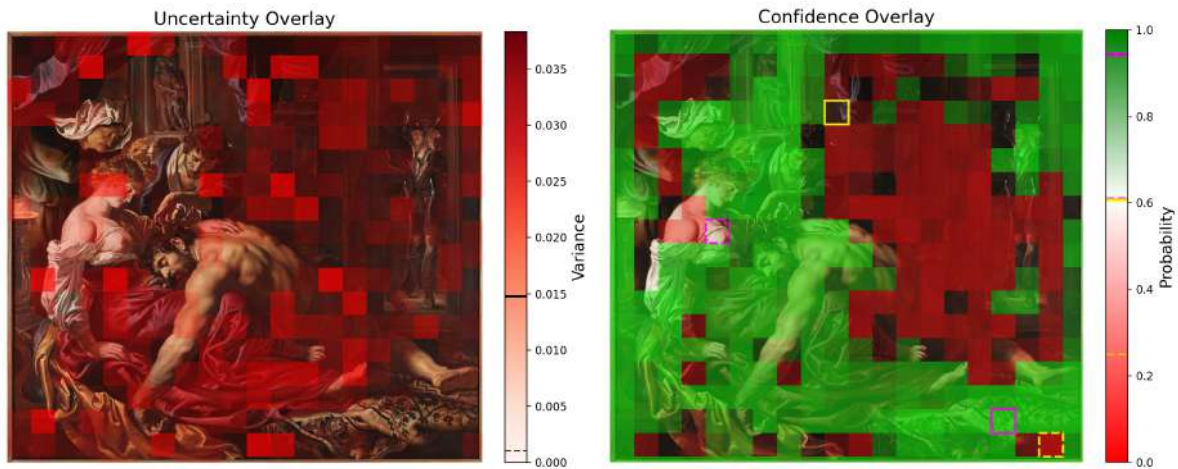

*Samson and Delilah* is a well-documented work by Rubens, but for a number of reasons outlined below, the model exhibited variability in its confidence. At the threshold of 0.6080, the average probability was 0.6547, and 255 out of 380 fragments were identified as consistent with the artist's style.

On the Uncertainty map, most high-variance red tiles are concentrated in darker, more uniform areas of the composition or in regions containing abrupt transitions, such as sharp boundaries between flesh tones and the surrounding background. These tiles tend to contain either low-detail textures or mixed visual cues that different models interpret inconsistently. In contrast, the figures themselves appear largely transparent, indicating very low disagreement across the ensemble. The human forms, such as faces, hands, musculature, provide rich, distinctive features that the networks classify in a consistent and stable way. This behavior indicates that the model's internal representation is strongly oriented toward the artist's characteristic handling of figures, rather than the more variable or less diagnostic elements of the composition.

The Confidence map indicates that the ensemble is generally confident across most of the central figures, although some green tiles appear slightly transparent, reflecting confidence levels that remain above the threshold but are not uniformly high across the entire surface. The right side of the composition and the background areas proved particularly unstable, with numerous red and nearly transparent zones. Many of these zones correspond closely to high-variance areas on the Uncertainty map, suggesting that regions provoking ensemble disagreement are also those where confidence drops.

Such a reaction can be partially explained by the complex conservation history of the painting. The restoration carried out by David Bomford in 1982 involved cleaning, retouching, and thinning of the wooden panel (Plesters, 1983). These interventions, along with localized paint losses, affected the surface texture, altering visual features crucial for the algorithm, such as brushstroke character, color balance, gloss, and microtexture.

In addition, a significant source of uncertainty lies in the stylistic heterogeneity of the painting itself. Created shortly after Rubens' return from Italy, the work incorporates several sources of inspiration. Samson's musculature alludes to classical sculpture, Delilah's pose recalls Michelangelo, and the lighting echoes Caravaggio's chiaroscuro. This stylistic eclecticism complicated the interpretation by a model trained primarily on Rubens' later, more stylistically consistent works.

This case demonstrates that even when a work is correctly classified, the model's final confidence may be reduced due to stylistic and textural variations. The algorithm is sensitive to deviations arising from both artistic experimentation and the passage of time. This underscores that, despite its overall effectiveness, machine analysis at this stage faces limitations when dealing with the layered nature of art, where stable visual features may lose clarity and require supplementation with historical and art historical context.

While we maintain our position in support of the painting's authenticity, it is worth noting that a small group of specialists - including, for example, the team at Art Recognition - has expressed reservations, or even outright rejection, citing similar technical ambiguities. Our interpretation, therefore, represents a different reading within the broader scholarly and analytical discourse.

*Charles I* (1635-before June 1636). Anthony Van Dyck.

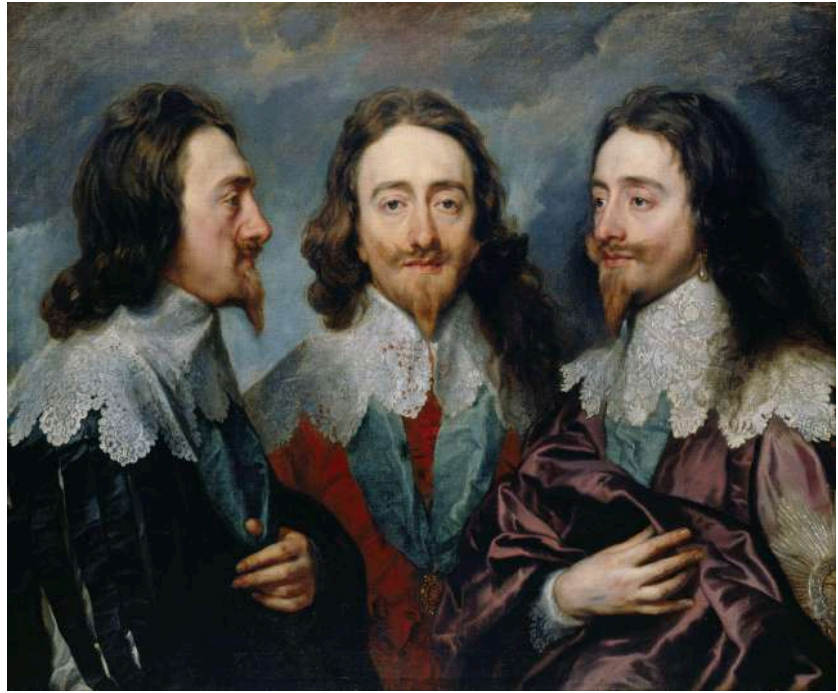

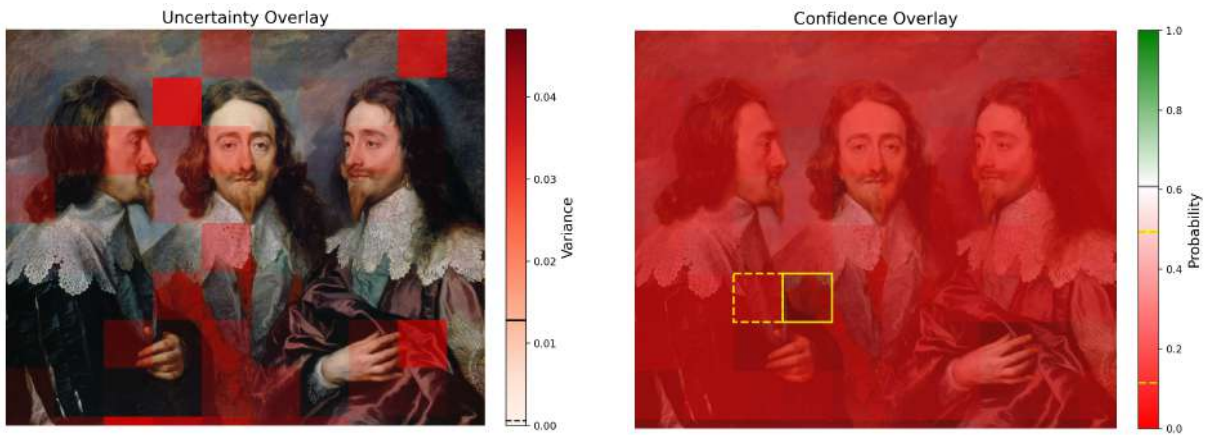

In this example, the Confidence map shows that all 90 tiles are rendered in red, indicating consistently low probabilities across the entire surface of the painting. The ensemble assigned an average confidence score of only 0.2883, well below the threshold of 0.6080, and no tile exceeded it. Taken together, these results mean that every fragment was classified as incompatible with Rubens' style and that the ensemble reached this conclusion with substantial certainty.

From an artistic standpoint, this uniformly low response indicates that the work lacks the characteristic features the algorithm, trained on Rubens' oeuvre, has internalised as markers of his authorship. The painting differs through its symmetrical composition with frontally posed figures — a structure uncharacteristic for Rubens — as well as refined brushwork lacking impasto and an overall absence of expressiveness, drama, and movement. These elements align poorly with the visual language the model expects from Rubens.

The context of the training data reinforces this interpretation. The algorithm was trained on a corpus of Rubens' works from the 1610s to the 1640s, dominated by multi-figure compositions, mythological and religious scenes, and a highly dynamic orchestration of motion, color, gesture, and dramaturgy. Against this backdrop, the painting in question appears stylistically marginal, and the consistently low scores across all tiles indicate that it provides few such cues. Even in visually intricate passages, such as faces or areas of abrupt tonal modulation, local computations remain within a uniformly low probability range, underscoring the absence of stylistic signals typically associated with Rubens.

The Uncertainty map provides extra detail. Although some tiles, particularly in complex regions, exhibit higher variance, these variations stay within a low-probability range and do not change the global classification. Despite minor internal variation, all models converge robustly toward rejecting the attribution to Rubens. Taken together, the evidence highlights the ensemble's ability to detect not only superficial visual similarities but also deeper structural and stylistic divergences, underscoring the tool's usefulness in attributional analysis.

*Head of a Young Man* (1601-1602)

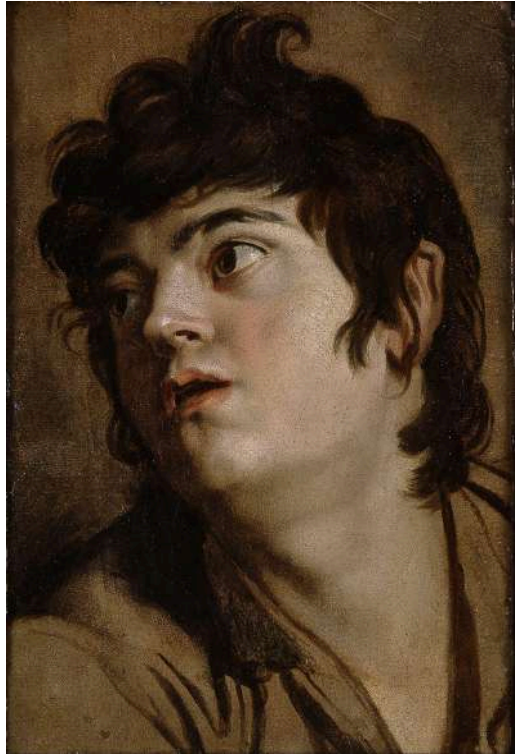

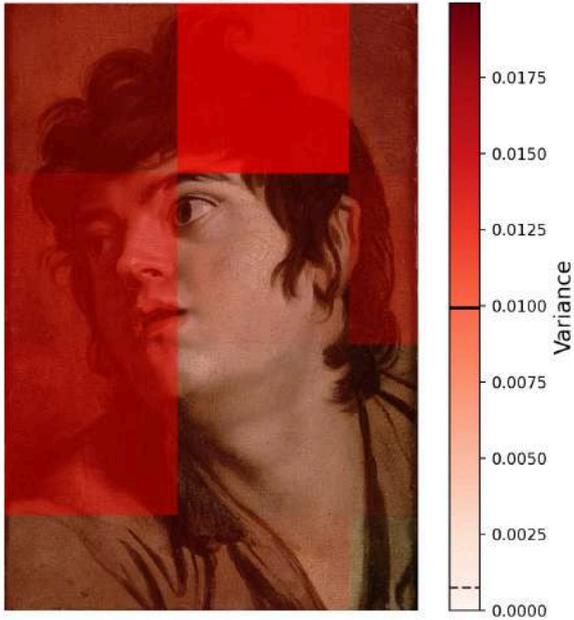
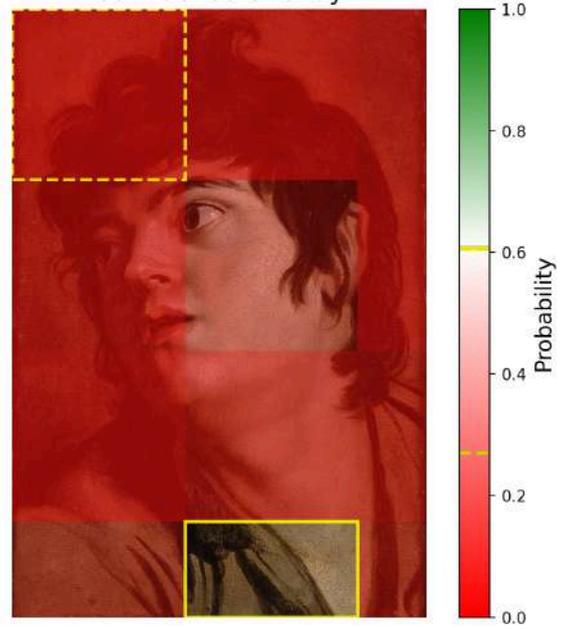

Although *Head of a Young Man* is recognized as an authentic work by Rubens, the model rejected it as non-authentic, as seen on the Confidence Map. The ensemble produced a low image-level confidence score (0.4310 compared with a threshold of 0.6080) and none of the tiles exceeded the attribution threshold. All image fragments were classified as inconsistent with the artist's style. The fragment enclosed by a solid yellow outline lies just below the threshold, providing only borderline support rather than a positive attribution. The most questionable areas were located in the upper and left part of the head and face, where the algorithm detected high variance, clearly visible on the Uncertainty map.

Part of this behavior can be attributed to the way the painting is tiled: many tiles capture only partial facial features or transitions between the head and the background, making it difficult for the ensemble to evaluate them confidently in isolation. Another likely factor is the structure of the training data. The dataset contains relatively few close-up head studies, so this painting occupies an underrepresented region of stylistic space, which further contributes to the model's uncertainty.

The painting's restrained composition — a single-figure academic study with subdued lighting and generalized modeling — stands in contrast to the dynamic, multi-figure works from Rubens' mature period (1610s–1640s), which formed the basis of the model's training data. As such, the algorithm did not detect the stylistic features it was trained to recognize and flagged the work as inconsistent. Notably, the attribution has been a subject of debate among art historians, including Julius Held, who questioned its place within Rubens' canon (Sutton, 2004).

Nonetheless, our research team does not question the painting's authenticity. This work is well-documented and widely accepted within the scholarly community. The model's negative classification reflects its narrow training scope rather than any conclusive statement about authorship. From our perspective, its output should be interpreted as highlighting areas of stylistic deviation, not as challenging established expert consensus.

This case illustrates how the model operates strictly within the limits of visual pattern recognition. Lacking awareness of broader historical context, stylistic evolution, or genre-specific characteristics, it treats deviations from its internalized visual norm as anomalies. It is, therefore, a tool to support — not replace — traditional connoisseurship.

Through the analysis of genuine Rubens paintings, the model has demonstrated its capacity to detect stylistic consistency within its training framework. At the same time, its limitations call for both cautious interpretation and further development, while underscoring the importance of integrating computational methods with expert art-historical judgment — a theme explored in the final section, *Discussion and Conclusions*.

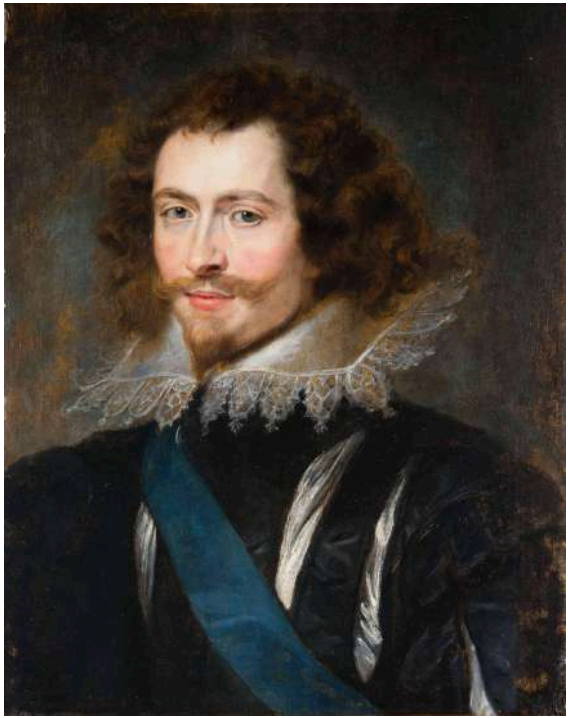

***Portrait of George Villiers, First Duke of Buckingham***
**Date unknown**
**Studio of Rubens**
Palazzo Pitti, Florence, Italy (whereabouts of original work unknown; presumably lost).

A particularly interesting case is the *Portrait of George Villiers, First Duke of Buckingham*. The original prototype of this portrait was recorded in Rubens' inventory, yet the work itself has not survived. With high probability, the present version represents the only known example of this type of portrait, executed as an exact copy based on the preparatory drawing of 1625, now held in the Albertina, Vienna.

From a technical standpoint, the ensemble assigned this work an average probability of 0.3410, significantly below the attribution threshold of 0.6080, indicating a strong and consistent assessment that the painting is not by Rubens. All tiles were classified as inconsistent with Rubens' style, with only two fragments falling slightly above the threshold. These two borderline areas, outlined on the visualization with solid and a dashed purple frame (marking the highest and lowest tile-level probabilities, respectively), are both located on the figure's face. Their placement suggests that the studio painter may have attempted to refine this key, visually expressive area more carefully than the rest of the composition.

The Uncertainty map demonstrates a comparatively low level of variability within the ensemble of models, which indicates strong consistency of the results: the system did not exhibit significant fluctuations in its judgments. The most noticeable zones of divergence are confined to the areas of the face and collar, which may be related to the studio painter's attempt to approximate the master's more refined and complex manner. In all other respects, the visualized data point to an overall stylistic homogeneity, distinct from Rubens' individual hand.

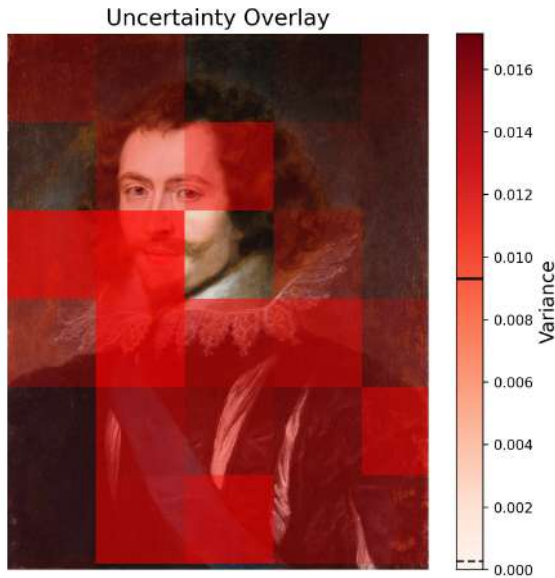

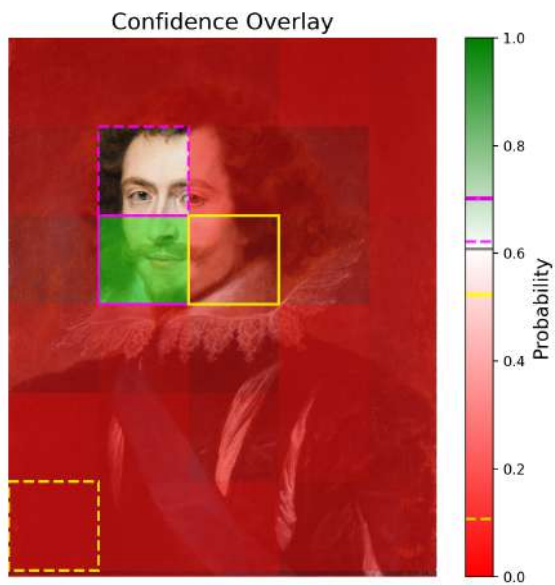

The Confidence map confirms this conclusion even more clearly. The overwhelming majority of fragments are rendered in deep red shades, which within the adopted methodology are interpreted as an extremely low probability of Rubens' participation. Even in those areas where the authorial hand is usually most evident, such as the face, the hands, and the costume details, the system did not detect features characteristic of Rubens' style. The absence of green and neutral zones reinforces this conclusion: the portrait is, with high probability, a studio copy.

These results are consistent with historiographical data: it is assumed that the work was executed in the workshop on the basis of the lost original, and that the copyist's task was not so much artistic interpretation as the careful reproduction of the iconographic prototype. Here the algorithm detected only extremely weak and unconvincing signals in the area of the face, which cannot be interpreted as signs of Rubens' hand.

Thus, the *Portrait of George Villiers, First Duke of Buckingham* serves as an illustration of how the algorithm makes it possible to distinguish clearly between autograph works or partially autograph ones and purely studio copies. In this case, the model's confident rejection of authorship not only supports the hypothesis of Rubens' non-involvement but also demonstrates that copies are characterized by stylistic homogeneity, markedly different from the more complex structure of works in which the master and his pupils collaborated.

As a result, it becomes evident that the proposed approach plays an important role not only in confirming or disproving hypotheses of authorship but also in the more detailed analysis of a work's internal structure. The algorithm is capable of localizing zones with higher probability of Rubens' participation and comparing them with those executed by his closest collaborators. In this way, it helps to refine the boundaries between the master's contribution and the workshop's production, making visible the very dynamics of interaction between Rubens and his circle. Such detailed analysis opens new perspectives for research and may guide art historians toward more substantiated conclusions in the attribution and reassessment of the role of the workshop in the creation of a work.

**VIII. Discussion and Conclusions**

From an art historical perspective, it is important to consider that between the original and the forgery exists a wide spectrum of intermediate cases: works involving pupils, late copies, restoration reconstructions. A telling example is the workshop of Rubens, where, due to its scale and the intensely collective nature of the work, the boundaries between an authored masterpiece and a pupil's production are often extremely blurred (Burckhardt, 1898). Our machine learning model is particularly useful for detecting visual inconsistencies that point to such scenarios, thereby guiding the expert's attention in the right direction.

In the future, the effectiveness of the system could be significantly enhanced by integrating into the analysis not only individual fragments but also the overall composition of the artwork, including its spatial organization and general structure. This would allow moving beyond microanalysis and taking into account the compositional features of the author's manner, thereby providing more accurate results.

Furthermore, a promising direction would be the inclusion of high-resolution images of the reverse side of works, which often contain key information about provenance. Their systematic scanning and interpretation could significantly improve the model's accuracy and contribute to the development of a more reliable automated attribution system.

Another important area of development is the expansion of machine learning methods to other forms of art, such as graphics, sculpture, or decorative and applied arts, where forgery is also relevant and where new approaches to expertise must be introduced.

Moreover, a significant step forward would be the development of algorithms capable of distinguishing complex methods of forgery, such as *pasticcio*, which involves compiling fragments from different authentic works by the same artist into a "new" stylistically convincing but essentially fake piece.

Another potential development vector for such systems could be functionalities that currently seem difficult to implement, for example, comparative analysis of X-ray images that would allow detecting hidden visual patterns characteristic of both forgeries and authentic works. It also seems promising to train the system to take into account not only the artist's signature style but also its evolution depending on the period, workshop, or even the physical and psycho-emotional state of the author. Such a system could rely on already documented and verified information collected by experts.

It is important to emphasize that this type of analysis should be seen as an important but auxiliary tool, since there is a risk of erroneous conclusions. For example, the results of machine learning based on a dataset of X-ray images may be contextually indeterminate, and in such cases it is advisable to involve specialists for qualified validation and interpretation. Furthermore, situations are possible where, for instance, an authentic signature is present on a work executed by a pupil, or when dealing with an especially skillful forgery that meticulously imitates the original.

Finally, in the future it would be appropriate to train machine analysis algorithms to be highly sensitive to the visual signs of the chemical-technological evolution of materials, for example, the characteristics of color, texture, or pigments that appeared only during specific historical periods. Even such approximate identification could help reveal discrepancies if the visually recognizable features do not align with the supposed dating or authorship.

In parallel with these long-term goals, the system has already demonstrated its effectiveness in several pilot applications. At this stage of development, our model should be understood primarily as a research and interpretive tool — one that does not replace traditional connoisseurship, but rather complements it by highlighting areas of stylistic ambiguity or inconsistency. In practice, the system has already demonstrated its utility when applied to a number of test cases. For example, the analysis of *Samson and Delilah* — a widely recognized work by Rubens — revealed stylistic and textural heterogeneity, likely influenced by both the painting's complex conservation history and its early, eclectic stylistic character.

In contrast, the *Head of a Young Man*, also attributed to Rubens, was flagged by the model as inconsistent with the artist's style, likely due to its academic format and the absence of features typical of Rubens' mature multi-figure compositions, on which the model was primarily trained.

These examples illustrate how the model can identify points of divergence from learned stylistic norms without making definitive claims about attribution. Its role is to raise questions, prompt closer examination, and provide new perspectives — especially in ambiguous cases where surface-level visual evidence may be inconclusive. In this sense, the tool fosters a more nuanced, multi-dimensional approach to attribution studies.

**Acknowledgements**

The authors would like to express sincere gratitude to Denis Moiseev for the opportunity to implement the project. Special thanks are extended to Konstantin Akinsha, whose invaluable methodological advice, professional guidance and moral support greatly contributed to this research; and to Maria Timina, for her insightful comments and encouragement.

This project was carried out entirely in the interests of Art Discovery.